\documentclass[sn-mathphys-num]{sn-jnl}

\usepackage{graphicx}%
\usepackage{multirow}%
\usepackage{amsmath,amssymb,amsfonts}%
\usepackage{amsthm}%
\usepackage{mathrsfs}%
\usepackage[title]{appendix}%
\usepackage{xcolor}%
\usepackage{textcomp}%
\usepackage{manyfoot}%
\usepackage{booktabs}%
\usepackage{algorithm}%
\usepackage{algorithmicx}%
\usepackage{algpseudocode}%
\usepackage{listings}%

\theoremstyle{thmstyleone}%
\theoremstyle{thmstyletwo}%

\theoremstyle{thmstylethree}%

\begin{document}

\title[Article Title]{\textbf{FacEnhance}: Facial Expression Enhancing with Recurrent DDPMs}


\author*[1]{\fnm{Hamza} \sur{Bouzid}}\email{hamza.bouzid@um5r.ac.ma}

\author[1]{\fnm{Lahoucine} \sur{Ballihi}}\email{lahoucine.ballihi@fsr.um5.ac.ma}

\affil[1]{\orgdiv{LRIT-CNRST URAC 29, Mohammed V University in Rabat}, \orgname{ Faculty Of Sciences}, \orgaddress{ \state{Rabat}, \country{Morocco}}}


\abstract{
Facial expressions, vital in non-verbal human communication, have found applications in various computer vision fields like virtual reality, gaming, and emotional AI assistants. Despite advancements, many facial expression generation models encounter challenges such as low resolution (e.g., 32x32 or 64x64 pixels), poor quality, and the absence of background details. In this paper, we introduce FacEnhance, a novel diffusion-based approach addressing constraints in existing low-resolution facial expression generation models. FacEnhance enhances low-resolution facial expression videos (64x64 pixels) to higher resolutions (192x192 pixels), incorporating background details and improving overall quality. Leveraging conditional denoising within a diffusion framework, guided by a background-free low-resolution video and a single neutral expression high-resolution image, FacEnhance generates a video incorporating the facial expression from the low-resolution video performed by the individual and with background from the neutral image. By complementing lightweight low-resolution models, FacEnhance strikes a balance between computational efficiency and desirable image resolution and quality. Extensive experiments on the MUG facial expression database demonstrate the efficacy of FacEnhance in enhancing low-resolution model outputs to state-of-the-art quality while preserving content and identity consistency. FacEnhance represents significant progress towards resource-efficient, high-fidelity facial expression generation, Renewing outdated low-resolution methods to up-to-date standards.
}

\keywords{Facial expression generation, Diffusion models, Face enhancing, FacEnhance, Generative}



\maketitle

\section{Introduction}\label{introduction}
Non-verbal human communication, encompassing facial expressions and human actions, holds significant importance in interpersonal interactions, with deep foundations in psychology, sociology, and cognitive science. Extensive research has been conducted on the analysis of this communication form \cite{baltrusaitis2018openface, zhu2022convolutional, de2024facial, peng2020learning, bouzid2024spatr}, unlocking new possibilities across various application fields, including virtual reality, gaming, telemedicine, and emotional AI assistants. With the advent of generative models, static 2D facial expression generation has achieved significant success \cite{choi2018stargan, yan2019joint, kim2023dcface, akram2024us}, yet, it falls short of capturing the dynamic nature of facial expressions, which involve continuous, smooth movements, rather than a static pose. Dynamic facial expression generation is less studied due to incorporating the temporal dimension, necessitating a dual focus on spatial and temporal understanding, respecting spatio-temporal consistency, and learning identity preservation through all video frames.
\vspace{0.2cm}

Researchers have developed diverse strategies for facial generation, including transfer-based methods for expression transfer \cite{wei2016facial, qiao2018emotional, qiao2018geometry}, the use of linear or non-linear coefficients to manage the expression intensity and temporal evolution \cite{ding2018exprgan}, spatio-temporal representation learning for one-step video generation \cite{vondrick2016generating,wang2020imaginator,bouzid2022facial}, and motion space learning for generating facial expression frames \cite{otberdout2020dynamic}. However, initial tests highlighted common limitations, such as low-quality and low-resolution results ($32x32$ or  $64x64$ pixels), difficulties in background generation, leading to distorted or absent backgrounds, and a lack of essential details like hair, neck, and clothing in the synthesized videos. 
\vspace{0.2cm}

To address these limitations, recent approaches aim to utilize successful image generators like StyleGAN \cite{karras2019style} for facial expression generation. These methods \cite{li2022stylet2i, yang2019unconstrained, hang2023language} involve learning to manipulate the embedding space of StyleGAN to generate the facial images. However, they struggle with content consistency since the generation process is performed frame-independent, and they are unable to manipulate images to a point where they lie outside the domain or in regions that are less covered by the pre-trained model. 
Another group addressing the low resolution and quality issue is the diffusion-based models (DM) \cite{ho2020denoising}. These models have demonstrated remarkable efficacy in image synthesis \cite{li2022srdiff, rombach2022high, saharia2022photorealistic, zhang2023survey}, excelling in capturing complex data distributions by employing a series of diffusion refinement steps. Diffusion-based models have achieved interesting results in facial expression generation \cite{ho2022video, ni2023conditional}. However, training and inference on diffusion models in the context of video processing, involving spatial and temporal dimensions, can be computationally demanding and time-consuming. This is especially pronounced due to the use of 3D convolutions in video diffusion, which further escalates the computational requirements and memory consumption. Thus, the training and inference processes become time and resource-intensive.
\vspace{0.2cm}

In response to the persistent challenges associated with low resolution and quality in resource-efficient facial expression generation methods, this paper introduces a novel diffusion-based approach, namely \textbf{FacEnhance}, for enhancing low-resolution efficient methods videos to the state-of-the-art level. Specifically, the proposed model addresses deficiencies in efficient models, such as low resolution, low quality, and lack of background details. It operates on generated videos to accomplish facial super-resolution, quality enhancement, and the addition of background/details. Taking low-resolution videos (64x64) of an individual performing a facial expression and a higher-resolution neutral image (192x192) of the same individual as input, the model integrates facial expressions with background and identity details, yielding videos of (192x192) pixels that combine the facial expression from the low-resolution video with the identity details and background sourced from the higher-resolution image. We note that the proposed model can theoretically generate videos that potentially reach even higher resolutions. However, we limit our work to (192x192) resolutions due to constraints in computational power.
\vspace{0.2cm}

The proposed model complements existing low-resolution facial expression generation models by enhancing their results spatially. By applying our model to low-resolution generated videos, we achieve a balance between efficiency and quality. In this study, we utilize facial expression generation models introduced in \cite{vondrick2016generating, wang2020imaginator, otberdout2020dynamic, bouzid2022facial} to produce black background low-resolution facial expression videos. Our model then enhances these videos, resulting in (192x192) facial expression videos with improved quality and integrated background details. The synergy of both models ensures high-quality facial expression generation without excessive computational demands.
\vspace{0.2cm}

We extensively evaluate our model quantitatively and qualitatively on the widely recognized MUG facial expression database. Our comparison includes recent state-of-the-art approaches such as LFDM \cite{ni2023conditional}, LDM \cite{rombach2022high}, VDM \cite{ho2022video}, and ImaGINator \cite{wang2020imaginator}. The results of our experiments demonstrate the effectiveness of our approach in enhancing low-resolution facial videos to state-of-the-art quality while maintaining content and identity consistency.

\section{Related Work}\label{relatedWork}

In the landscape of facial expression generation, the pursuit of high-quality results with efficient computational methods has inspired diverse approaches. Some methods, known as transfer-based methods, focus on transferring expression from one video to another, either using embeddings \cite{wiles2018x2face}, action units \cite{ling2020toward}, or facial landmarks \cite{qiao2018emotional, qiao2018geometry, song2018geometry, siarohin2019animating} extracted from the frames of the target video. These methods do not learn motion dynamics but extract expressions from other videos, relying on the quality and diversity of source data. Additionally, they often operate frame-by-frame, resulting in less natural and temporally inconsistent outcomes. Another kind of approaches \cite{ding2018exprgan, kang2023gammagan} involves using linear coefficients to control the intensity and temporal evolution of expressions, making them dynamic. Other approaches use an interpolation technique by blending between key expressions. However, these two methods generate oversimplified expressions with linear or unrealistic temporal evolution. 
Another group decomposes the video into content and motion information through two different streams to control them separately \cite{tulyakov2018mocogan, sun2020twostreamvan}. Nevertheless, results obtained from these models showed artifacts and noise. An alternative group seeks to generate videos directly by learning the spatio-temporal representation of facial expressions \cite{vondrick2016generating, wang2020imaginator, wang2020g3an, bouzid2022facial}. While promising, this method struggles with the high complexity, computations, and resource consumption, which limits the generated images to very low resolution. 
Another group of methods \cite{otberdout2020dynamic} aims to learn a motion space and sample from it to generate motion embedding. The motion embedding is then used by an image-to-image generator to synthesize the facial expression frames. However, this group of methods falls short when handling the spatio-temporal consistency, leading to clear images individually but videos with noises and artifacts. Alternative approaches have explored the potential of using recurrent models in facial expression generation \cite{Gupta2022RV, Zhu2020S3VAESS}. The majority of the methods discussed above focus on generating facial expressions within a limited scope, typically isolating the face area without the hair, neck region, or background. Furthermore, many of these approaches often produce expressions at low resolutions, typically 64x64 pixels. Addressing these issues, researchers have explored adapting successful image generators, such as StyleGAN \cite{karras2019style}, for facial expression generation. This involves learning to manipulate the embedding space of the pre-trained StyleGAN to generate video frames. StyleGAN-based methods \cite{zhang2021facial, burkov2020neural, li2022stylet2i, yang2019unconstrained, hang2023language} excel in achieving high-resolution and quality, but struggle with maintaining the content-consistency in dynamic expressions, leading to unnatural visual transitions in the in generated videos. They also face challenges when manipulating images beyond their pre-trained domain. In addition, fine-grained control over specific facial features is limited as StyleGAN was designed for high-level image synthesis rather than precise expression manipulation. Another group addressing the low resolution and quality issue is the diffusion-based models. These models stand as the current state-of-the-art in class-conditional image synthesis. Video diffusion models (VDM) \cite{ho2022video} use a combination of diffusion and 3D convolutions to generate facial expression videos, leading to high content consistency. However, the task complexity and the computational demands of diffusion models in the video domain are significant, compared to their 2D counterparts. This imposes constraints on resolution and frame count (e.g., (16x64x64) to (9x128x128)), despite substantial resources, and compromises on details and the production of blurry, low-quality videos. In contrast to Video Diffusion Models (VDM), which couples spatial and temporal aspects in diffusion synthesis, latent flow diffusion model (LFDM) \cite{ni2023conditional} separates the task of facial expression generation into distinct spatial and temporal components. More precisely, LFDM employs a 3D-UNet-based diffusion model to synthesize optical flow sequences representing the motion in the latent space. These generated optical flow sequences are then utilized by a conditional auto-encoder to animate the input image.
\vspace{0.2cm}

Motivated by the discussion above, we present, in this paper, a diffusion-based model for facial expression super-resolution and enhancement. The rest of this paper is structured as follows. Section \ref{preliminaries} provides an overview of diffusion models. In Section \ref{proposedApproach}, we present the novel FacEnhance model. Section \ref{exprmt} presents the experimental configurations, encompassing both quantitative and qualitative analyses. The paper concludes in Section \ref{conclusion}, offering insights into future research directions.

\section{PRELIMINARIES: DIFFUSION MODELS}
\label{preliminaries}

\begin{figure*}[ht]
\begin{center}
\includegraphics[scale=0.4]{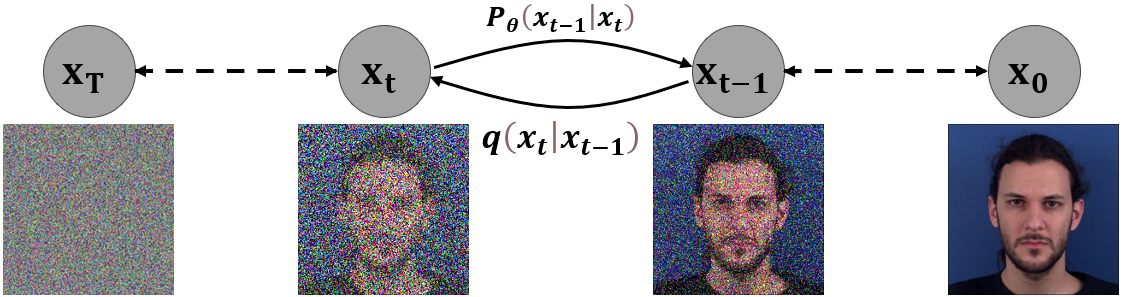}
\caption{A graphical representation of diffusion models, highlighting the noise diffusion process $q(x_t|x_{t - 1})$ and the denoising process $p_{\theta}(x_{t - 1}|x_t)$}
\label{diff}
\end{center}
\end{figure*}

Diffusion models, introduced in "Denoising Diffusion Probabilistic Models" \cite{ho2020denoising}, revolutionize image generation by framing it as a denoising process. Instead of directly sampling from a probability distribution, DDPM refines a noisy image iteratively through denoising steps, implicitly learning the underlying data distribution for high-quality sample generation. DDPM comprises two key processes: the \textbf{diffusion process} $q(x_t | x_{t-1})$ and the \textbf{reverse process} $p_\theta(x_{t-1} | x_t)$. Example illustrated in Fig.\ref{diff}.

\textbf{Diffusion Process:} maps the initial data distribution $(q(x_0))$ to the latent variable distribution $(q(x_T)\sim \textbf{$\mathcal{N}$(0, I)}$) through a Markov chain with Gaussian noise (\(\varepsilon\)). Mathematically, it is expressed as:

\vspace{-0.5cm}
\begin{equation*} \label{eq1}
\begin{split}
q(x_1, \ldots, x_T | x_0) & := \prod_{t=1}^{T} q(x_t | x_{t-1}), \\
q(x_t | x_{t-1}) & := \mathcal{N}\left(x_t; \sqrt{1 - \beta_t} x_{t-1}, \beta_t \textbf{I}\right),
\end{split}
\end{equation*}

\noindent where $\beta_t$ is a small positive constant hyperparameter controlling noise amplitude. Defining $\alpha_t := 1 - \beta_t$ and $\bar{\alpha}_t := \prod_{s=1}^{t} \alpha_s$, $\bar{\alpha}_t$ represents the cumulative noise diffusion complement up to the current time step. This facilitates closed-form sampling $x_t$ at any arbitrary time step $t$, leading to enhanced training efficiency.

\begin{equation} \label{eq2}
q(x_t | x_0) = \mathcal{N}\left(x_t; \sqrt{\bar{\alpha}_t} x_0, (1 - \bar{\alpha}_t)\textbf{I}\right)
\end{equation}

\begin{equation} \label{eq3}
x_t(x_0, \varepsilon) = \sqrt{\bar{\alpha}_t} x_0 + \sqrt{1 - \bar{\alpha}_t}\varepsilon, \quad \varepsilon \sim \textbf{$\mathcal{N}$(0, I)}
\end{equation}

\textbf{Reverse Process:} applying Bayes theorem, the authors established the Gaussian nature of the posterior distribution ($q(x_{t-1} | x_t, x_0)$) \cite{ho2020denoising}:

\begin{equation}\label{eq7}
    q(x_{t-1}|x_t, x_0) = \mathcal{N}(x_{t-1}; \widetilde{\mu}_t(x_t, x_0), \widetilde{\beta}_tI),
\end{equation}

\noindent with $\widetilde{\mu}_t(x_t, x_0)$ and $\widetilde{\beta}_t$ defined to approximate ($q(x_{t-1}|x_t, x_0)$). An approximation of ($q(x_{t-1}|x_t, x_0)$) eq.\ref{eq7} is proposed to map the latent variable distribution ($p_\theta(x_T)\sim \textbf{$\mathcal{N}$(0, I)}$) to the data distribution (\(p_\theta(x_0)\)), where \(\theta\) represents the learnable parameters. This mapping is characterized as a Markov chain employing Gaussian transitions, initiated with \(p(x_T) = \mathcal{N}(x_T; 0, I)\):

\vspace{-0.6cm}
\begin{equation} \label{eq4}
\begin{split}
p_\theta(x_0, \ldots, x_{T-1} | x_T) &:= \prod_{t=1}^{T} p_\theta(x_{t-1} | x_t), \\
p_\theta(x_{t-1} | x_t) &:= \mathcal{N}\left(x_{t-1}; \mu_\theta(x_t, t), \sigma_\theta(x_t, t)^2\textbf{I}\right)
\end{split}
\end{equation}

\section{Proposed Approach}\label{proposedApproach}
\begin{figure*}[t]
\begin{center}
\includegraphics[scale=0.55]{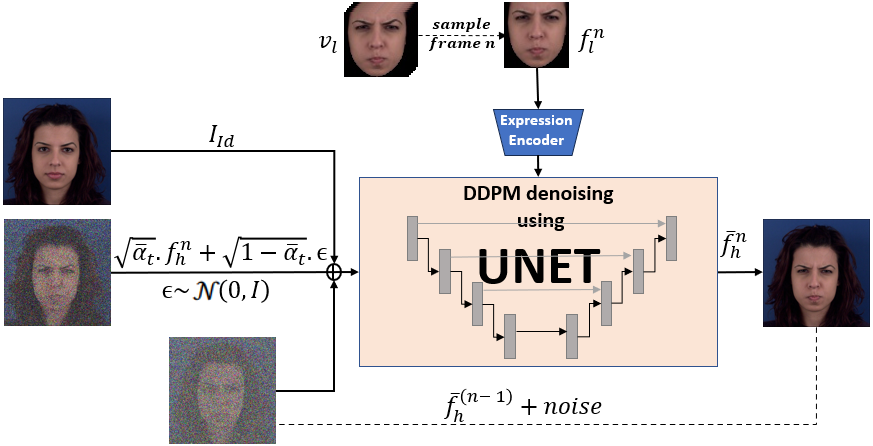}
\caption{Overview of the proposed facial expression enhancement model. The diffusion model refines Gaussian noise iteratively, guided by a low-resolution frame \(f_{l}^{n}\) from input video \(v_{l}\), a neutral high-resolution image \(I_{Id}\), and the previously generated frame \(\Bar{f}_{h}^{n-1}\), resulting in an improved higher-resolution frame \(\Bar{f}_{high}^{n}\).}\label{ourApproach}
\end{center}
\end{figure*} 

The primary objective of our model is to enhance the quality, resolution, and background of facial expression videos generated by low-resolution models. The approach involves processing a low-resolution (64x64) facial expression video, in conjunction with a higher-resolution input image containing the same person with a neutral facial expression, background, and additional details. The objective is to generate an improved higher-resolution (192x192) video, featuring enhanced facial expressions, background, and finer details. This task is framed as training a model to learn a function denoted as: \(\text{FacEnhance}(v_{low}, I_{Id}) \rightarrow \Bar{v}_{high}\), where \(v_{low} = [f_{low}^{0},f_{low}^{1},...,f_{low}^{N}]\) represents the low-resolution video with \(N\) frames, \(I_{Id}\) is the higher-resolution input image, and \(\Bar{v}_{high} = [\Bar{f}_{high}^{0},\Bar{f}_{high}^{1},...,\Bar{f}_{high}^{N}]\) signifies the generated higher-resolution video of \(N\) frames. 
\vspace{0.2cm}

To elaborate, each $n^{th}$ low-resolution frame \(f_{low}^{n}\) (sampled from \(v_{low}\)), combined with the neutral input image \(I_{Id}\) and the high-resolution frame generated from the previous step \(\Bar{f}_{high}^{n-1}\), is fed into a diffusion model. The diffusion model then generates the high-resolution frame \(\Bar{f}_{high}^{n}\) incorporating background details, expressed as \(\text{Diff}(f_{low}^{n}, I_{Id}, \Bar{f}_{high}^{n-1}) \rightarrow \Bar{f}_{high}^{n}\). Finally, all frames are aggregated to construct the complete high-resolution video \(\Bar{v}_{high}\). Refer to Fig.\ref{ourApproach} for an overview of the proposed model.

\subsection{Our proposed Model: FacEnahance} 
In our proposed model, we use conditional denoising to guide the facial image generation. The condition label is composed of 3 guiding images: 1) $n^{th}$ low-resolution frame $f_{low}^n$, that we use to inject the expression, 2) identity image $I_{Id}$, for identity details and background, and 3) previous generated high-resolution frame $\Bar{f}_{high}^{n-1}$, which is used for the content consistency preservation. The conditional distribution $p_{\theta}(x|I_{Id}, f_{low}^n, \Bar{f}_{high}^{n-1} )$ enables the generation of data dependent on the specified conditioning images.
\vspace{0.2cm}

Throughout the diffusion process, noise is systematically injected into the input image, independent of the conditioning label $q(\Bar{f}^n_{high, t} | \Bar{f}^{n}_{high, t-1})$. On the other hand, the denoising model $p_{\theta}(\Bar{f}^{n}_{high, t-1} | \Bar{f}^n_{high, t}, I_{Id}, f_{low}^n, \Bar{f}_{high}^{n-1})$ considers both the image from the noisy image (\(\Bar{f}^n_{high, t}\)) and the conditioning labels (\(I_{Id}, f_{low}^n, \Bar{f}_{high}^{n-1}\)).
\vspace{-0.2cm}
\begin{multline*} \label{cond_reverse}
p_\theta(\Bar{f}^n_{high, 0}, \ldots, \Bar{f}^n_{high, T-1} | \Bar{f}^n_{high, T}, I_{Id}, f_{low}^n, \Bar{f}_{high}^{n-1}) := \\ := \prod_{t=1}^{T} p_\theta(\Bar{f}^n_{high, t-1} | \Bar{f}^n_{high, t}, I_{Id}, f_{low}^n, \Bar{f}_{high}^{n-1}),
\end{multline*}
\vspace{-.3cm}
with 
\begin{multline*}
    p_\theta(\Bar{f}^n_{high, t-1} | \Bar{f}^n_{high, t}, I_{Id}, f_{low}^n, \Bar{f}_{high}^{n-1}) := \\
    := \mathcal{N} (\Bar{f}^n_{high, t-1}; \mu_\theta(\Bar{f}^n_{high, t}, t, I_{Id}, f_{low}^n, \Bar{f}_{high}^{n-1}), \sigma_\theta(\Bar{f}^n_{high, t}, t, I_{Id}, f_{low}^n, \Bar{f}_{high}^{n-1})^2\textbf{I} ).
\end{multline*}

\subsubsection{Inputs} The choice of ($I_{Id}, f_{low}^n, \Bar{f}_{high}^{n-1}$) as guidance to the denoising process is based on the information available in each one of the images. $f_{low}^n$, which is the frame sampled from the low-resolution video, contains information about the performed expression. $I_{Id}$, the high-resolution input image of the same person performing a neutral expression, contains the facial identity and background information. $\Bar{f}_{high}^{n-1}$, the previously generated frame, creates a temporal dependency between frame generation, ensuring that the generated facial frames maintain temporal consistency, an essential characteristic for video content generation. Removing this frame causes independently generated frames to be slightly different, which would cause distortions in the form of vibrant noise in the video. Initial attempts with direct injection of the $\Bar{f}_{high}^{n-1}$ yielded inferior results, as the model overly relied on specific details of $f_{high}^{n-1}$ in the training data (real frame). During the inference phase, where the generated $\Bar{f}_{high}^{n-1}$ frame is utilized, there is a cumulative loss of information and noise across all preceding frames generation processes, intensifying the distortions observed in the generated frames. To handle this, we intentionally add noise during training to obscure facial details, creating a ($\Bar{f}_{high}^{n-1}+z$) frame where only the silhouette is distinguishable. This step eliminates dependency on $\Bar{f}_{high}^{n-1}$ details, retaining only information about head, hair, and components positions. The inputs $I_{Id}$ and $\Bar{f}_{high}^{n-1}+z$ are concatenated together and combined with the noised image $\Bar{f}_{high, T}^{n}$, and $f_{low}^n$ is encoded using a proposed expression encoder $E_{exp}$ and injected in the embedding space of the model. The effects of $f_{low}^n$ injection technique, $\Bar{f}_{high}^{n-1}$ and the added noise $z$ are analyzed in section \ref{sec:ablation}.

\subsubsection{The used architecture} Following the prior works on diffusion models, we employ a conditional-UNET architecture for denoising. Noteworthy features include attention blocks for emphasizing facial features, residual blocks for facilitating information flow, and temporal embedding for considering noise levels at specific timesteps. Our model directly predicts the denoised image $\Bar{f}_{high, t-1}^{n}$ from the input, which includes four RGB images (noisy $\Bar{f}_{high, t}^n, I_{Id}, f_{low}^n, \Bar{f}_{high}^{n-1}$), and the diffusion timestep $t$. 
\vspace{0.2cm}

In addition, we introduce an extra encoder, $E_{\text{exp}}$, specifically designed to extract expression embeddings from the expression frames $f_{\text{low}}^n$. These features are then incorporated into the UNET embedding space via a self-attention layer. This self-attention mechanism serves as a selective information focus based on the unique characteristics of facial expressions and aims to enhance the contextual understanding of relationships within the input sequence. The objective of this modification is to utilize expression information to guide the denoising process. The impact of this encoder is highlighted in our ablation study (Section \ref{sec:ablation}).
\vspace{0.2cm}

Since our model predicts the denoised image $\Bar{f}_{high, t-1}^{n}$ considering expression information, training involves minimizing the Mean Squared Error (MSE) between the predicted denoised images and their ground-truth counterparts:

\begin{equation}\label{eq8}
\mathcal{L}_{MSE} = |\Bar{f}_{high, t-1}^{n} - f_{high, t-1}^{n}|^2
\end{equation}

This loss function captures the overall structure and enables preserving identity details, while incorporating expression-related features through the introduced encoder.

\subsection{Training}
Algorithm \ref{alg:training} outlines the training procedure for a Denoising Diffusion Probabilistic Model (DDPM) designed for generating high-resolution facial images from low-resolution inputs. The process involves \textbf{(1)} initializing a conditional denoiser (\( \theta \)) and \textbf{(2)} iteratively refining it through a training loop. Within each iteration, \textbf{(3)} samples from the dataset comprising low-resolution frames, high-resolution identity images, and previous high-resolution frames are drawn. \textbf{(4)} Gaussian noise is introduced, and a diffusion timestep is chosen. \textbf{(5,6)} The high-resolution frames undergo a diffusion process to generate $f^n_{high, t}$ and $f^n_{high, t-1}$, and \textbf{(7)} a denoising model (\( \theta \)) use $f^n_{high, t}$ to estimate the previous high-resolution frame $f^n_{high, t}$. \textbf{(8)} The training objective involves minimizing the difference between the estimated frame and the ground truth. This iterative procedure continues until convergence \textbf{(9)}, resulting in a denoiser capable of generating high-quality facial images.

\begin{algorithm}
\caption{Training Algorithm}\label{alg:training}
\begin{algorithmic}[1]
\Require Low-resolution frames (n), their corresponding high-resolution frames (n), high-resolution identity images, and high-resolution previous frames (n-1); $ D~=~\{(f^n_{low}, I_{Id}, f^n_{high}, f^{n-1}_{high})\}^{K}_{k=1}$, total steps~$T$. 
\State \textbf{Initialize:} randomly initialized conditional denoiser $\theta$ \hspace*{\fill}\linebreak \Comment{here we consider $\theta$ as the entire network containing both the UNET and $E_{exp}$}
\Repeat
\State Sample $(f^n_{low}, I_{Id}, f^n_{high}, f^{n-1}_{high}) \sim D$
\State Sample $\varepsilon \sim \mathcal{N}(0, I)$, and $t \sim \text{Uniform}(\{1, \ldots, T\})$
\State $f^n_{high, t} \gets \sqrt{\bar{\alpha}_t} f^n_{high} + \sqrt{1 -\bar{\alpha}_t}\varepsilon$
\State $f^n_{high, t-1} \gets \sqrt{\bar{\alpha}_{t-1}} f^n_{high} + \sqrt{1 -\bar{\alpha}_{t-1}}\varepsilon$
\State $\Bar{f}^n_{high, t-1} \gets \theta(f^n_{high, t}, I_{Id}, f^n_{low}, f^{n-1}_{high}+z)$ \Comment{$z \sim \mathcal{N}(0, I)$}
\State Take gradient step on $\nabla_{\theta}|| \Bar{f}^n_{high, t-1} - f^n_{high, t-1} ||_2$
\Until{convergence}
\end{algorithmic}
\end{algorithm}

\subsection{Inference}
Algorithm \ref{alg:inference} describes the frame inference process that we follow in this work. Given a low-resolution frame \(f^n_{low}\), a high-resolution identity image \(I_{Id}\), and the previous high-resolution frame \(f^{n-1}_{high}\), \textbf{(1)} the algorithm employs a conditional denoiser \(\theta\) to infer the high-resolution frame \(f^n_{high, 0}\). \textbf{(2)} The procedure involves sampling from a normal distribution and \textbf{(3)} iteratively applying the denoiser in reverse diffusion steps (\(t = T, T-1, ..., 1\)). At each step, \textbf{(4)} the denoiser refines the estimate of the previous high-resolution frame \(\Bar{f}^n_{high, t-1}\). \textbf{(6)} The final output is the inferred high-resolution frame \(\Bar{f}^n_{high, 0}\), representing a denoised version of the input frame. 

\begin{algorithm}
\caption{Frame Inference Algorithm}\label{alg:inference}
\begin{algorithmic}[1]
\Require Low-resolution frame $f^n_{low}$, high-resolution identity image $I_{Id}$, high-resolution previous frame $f^{n-1}_{high}$, total diffusion steps $T$. 
\State \textbf{Load:} conditional denoiser $\theta$
\hspace*{\fill}\linebreak \Comment{here we consider $\theta$ as the entire network containing both the UNET and $E_{exp}$}
\State sample $f^n_{high, T} \sim \mathcal{N}(0, I)$
\For{t = T, T-1, ..., 1}
\State $\Bar{f}^n_{high, t-1} \gets \theta(f^n_{high, t}, I_{Id}, f^n_{low}, f^{n-1}_{high}+z)$ \Comment{$z \sim \mathcal{N}(0, I)$}
\EndFor
\State \textbf{return} $\Bar{f}^n_{high, 0}$
\end{algorithmic}
\end{algorithm}

\subsection{video Inference}
Algorithm \ref{alg:vidInference} outlines the video inference process within the framework of a Denoising Diffusion Probabilistic Model (DDPM). Given a low-resolution video \(v_{low} = \{(f^n_{low})\}_{n=1}^N\), a high-resolution identity image \(I_{Id}\), and the total diffusion steps \(T\), \textbf{(1)} the algorithm utilizes a conditional denoiser \(\theta\) to generate a high-resolution video. The iterative procedure involves \textbf{(2)} initializing the previous high-resolution frame \(f^{n-1}_{high}\) with the identity image. For each frame \(f^n_{low}\) in the low-resolution video, \textbf{(4)} the algorithm samples a high-resolution frame \(f^n_{high, T}\) from a normal distribution. Subsequently, \textbf{(5,6)} the denoiser is applied in reverse diffusion steps (\(t = T, T-1, ..., 1\)) to generate denoised frames \(\Bar{f}^n_{high, 0}\). \textbf{(8)} These denoised frames are accumulated to form the generated high-resolution video. The algorithm iterates through all frames in the low-resolution video, \textbf{(9)} updating the previous high-resolution frame at each step. This process showcases the model's capability to generate high-quality, denoised videos from low-resolution inputs.

\begin{algorithm}
\caption{Video Inference Algorithm}\label{alg:vidInference}
\begin{algorithmic}[1]
\Require Low-resolution video $v_{low} = \{(f^n_{low})\}_{n=1}^N$, high-resolution identity image $I_{Id}$, total diffusion steps $T$. 
\State \textbf{Load:} conditional denoiser $\theta$
\hspace*{\fill}\linebreak \Comment{here we consider $\theta$ as the entire network containing both the UNET and $E_{exp}$}
\State $f^{n-1}_{high} \gets I_{Id}$
\For{n = 1, 2, ..., N}
    \State sample $f^n_{high, T} \sim \mathcal{N}(0, I)$
    \For{t = T, T-1, ..., 1}
    \State $\Bar{f}^n_{high, t-1} \gets \theta(f^n_{high, t}, I_{Id}, f^n_{low}, f^{n-1}_{high}+z)$ \Comment{$z \sim \mathcal{N}(0, I)$}
    \EndFor
    \State Add $\Bar{f}^n_{high, 0}$ to the generated video
    \State $f^{n-1}_{high} \gets \Bar{f}^n_{high, 0}$
\EndFor
\end{algorithmic}
\end{algorithm}

\section{EXPERIMENT}
\label{exprmt}
To evaluate our model, we perform an extensive evaluation on the MUG database \cite{aifanti2010mug}, covering quantitative and qualitative analyses. We detail the experimental setup, learning process, and evaluation criteria. The model is evaluated on low-resolution videos (64x64) from the MUG dataset and on low-resolution videos generated by VideoGAN, ImaGINator, FEV-GAN, and by MotionGAN \cite{otberdout2020dynamic}. We compare our results with state-of-the-art models: ImaGINator \cite{vondrick2016generating}, VDM\cite{ho2022video}, LDM \cite{rombach2022high}, and LFDM\cite{ni2023conditional}.
\subsection{DATASET}
To train and evaluate our model, we have used the Multimedia Understanding Group (\textbf{MUG}) database \cite{aifanti2010mug}. This database comprises videos featuring 86 individuals (52 subjects available to authorized Internet users, 25 subjects accessible upon request, and the remaining 9 subjects exclusively accessible within the MUG laboratory) expressing seven distinct emotions: "happiness," "sadness," "surprise," "anger," "disgust," "fear," and "neutral". Notably, each video starts and ends with a neutral expression, with the apex expression displayed in the middle. The videos consist of 50 to 160 RGB frames, each with a resolution of 896×896 pixels. For our evaluation, we utilize the public data of 52 subjects, specifically focusing on the first half (neutral to apex expression) of the videos representing the six basic expressions.

\subsection{IMPLEMENTATION DETAILS}
For experimentation, we perform a subject-independent split, allocating $75\%$ for learning and $25\%$ for testing. Videos are standardized to N=32 frames, each frame is resized to (192x192) to create the high-resolution videos $v_{high}$. OpenFace \cite{baltrusaitis2018openface} is used to extract low-resolution videos $v_{low}$ with facial expressions against a black background (32x64x64). 
\vspace{0.2cm}

The input data is assumed to have 9 channels (noisy image $f^n_{high}$, Identity image $I_{Id}$, and the previously generated frame $\Bar{f}^{n-1}_{high,t}$). The model incorporates 128 hidden channels. Channel multipliers vary across different levels, following the configuration [1, 1, 2, 2, 4, 8], indicating the number of times the hidden channels are multiplied at each level. The model comprises one residual block per level, with attention blocks selectively applied, specifically at the fifth level, where embedded expression information $E_{exp}(f^n_{low})$ is injected into the network. The diffusion process unfolds over 1000 timesteps, with beta values ranging from 0.0001 to 0.02 in a linear schedule. 
\vspace{0.2cm}

During training, the model is trained for $400$ epochs, reaching about 2 million steps, with a learning rate of $2 \times 10^{-5}$ and a batch size of 4. Exponential Moving Average (EMA) is utilized, with a decay rate of 0.9999. The training is performed on a single Nvidia Titan V GPU (12GB of memory) for about 140 hours.
\vspace{0.2cm}

To evaluate our model against videos generated by low-resolution techniques, we create facial expression videos using \cite{vondrick2016generating, bouzid2022facial, otberdout2020dynamic}. For benchmarking against state-of-the-art models, we employed public codes of ImaGINator \cite{wang2020imaginator} and LFDM \cite{ni2023conditional} making minor modifications for conditionality and resolution. Results for VDM \cite{ho2022video} and FDM \cite{rombach2022high} were obtained from \cite{ni2023conditional}.

\subsection{EVALUATION METRICS}
In our quantitative assessment, we employ various metrics for a comprehensive evaluation:
\begin{itemize}
    \item \textbf{FVD $\downarrow$} (Frechet Video Distance): measures the dissimilarity between the distributions of features extracted from generated and real videos using a pre-trained classifier.
    \item \textbf{PSNR $\uparrow$} (Peak Signal-to-Noise Ratio): Measures pixel-level similarity between generated videos and their ground truth.
    \item \textbf{SSIM $\uparrow$} (Structural Similarity Index Measure): Quantifies the structural similarity between real and reconstructed videos.
    \item \textbf{ACD $\downarrow$} (Average Content Distance): Evaluates facial identity consistency by calculating average pairwise L2 distances between facial features of consecutive frames. However, it focuses solely on identity consistency within the video.
    \item \textbf{ACD-I $\downarrow$} (Average Content Distance - Identity): An extension of ACD, focusing preservation of the input identity in the generated video. It calculates average L2 distances between the facial features of generated video frames and the input image.
\end{itemize}

\subsection{EXPERIMENTAL RESULTS}
\subsection{Facial Expression Enhancing}
In the process of validating the efficacy of our model in enhancing facial videos, we produce videos showcasing 13 distinct individuals, from the testing set, performing the six basic facial expressions. Our evaluation involves observing the effect of our model on videos from various models, namely VideoGAN, ImaGINator, MotionGAN, and FEVGAN. We specifically examine the quality both before and after applying enhancements by our model to understand the impact of our proposed methodologies. We note that our model is trained on the original dataset and has not processed any generated data or the testing identities during training.
\vspace{0.2cm}

\begin{figure}[t]
\begin{center}
\includegraphics[scale=0.48]{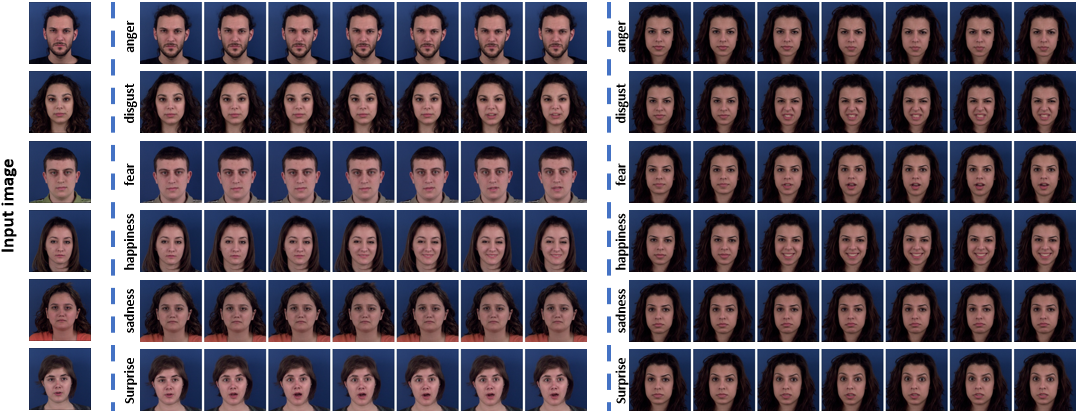}
\caption{Video examples generated by our model showcasing the six basic facial expressions.The right side features generated videos of the six facial expressions performed by the same individual, while the left side presents six different individuals, with each person expressing one specific emotion.}
\label{example-gen}
\end{center}
\end{figure}

\begin{figure}[t]
\begin{center}
\includegraphics[scale=0.5]{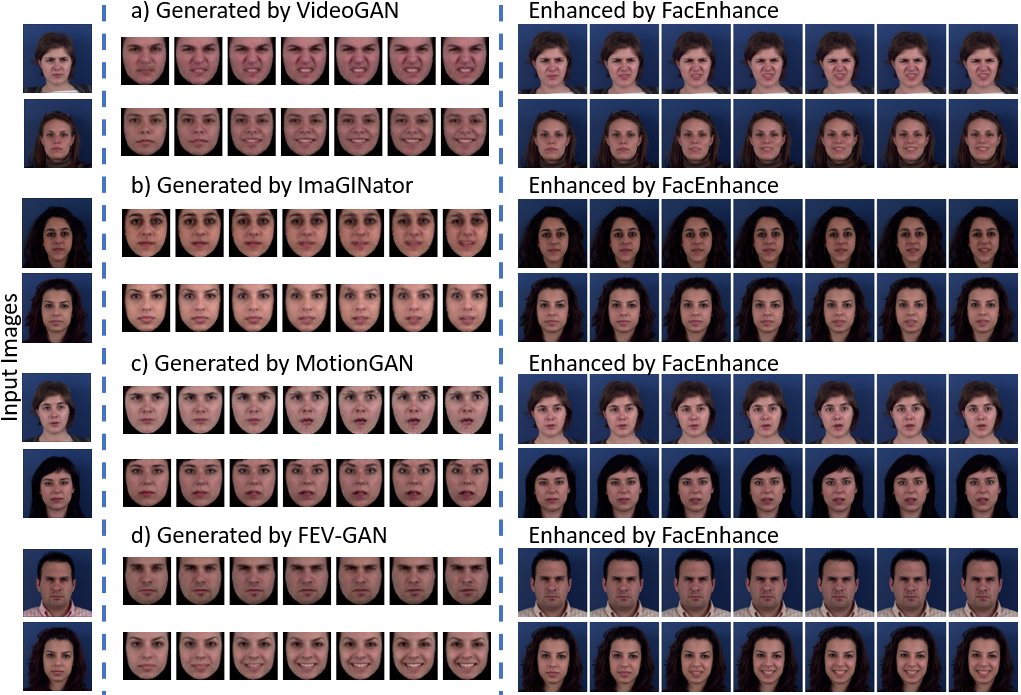}
\caption{Qualitative comparison of facial expression sequences before and after enhancement using our model. On the far left of the figure, we display the input images. Adjacent to them are videos generated by VideoGAN (a), ImaGINator (b), MotionGAN (c), and FEV-GAN (d). On the right side, we present the corresponding enhanced videos by FacEnhance.}
\label{example-enh}
\end{center}
\end{figure}

The evaluation is conducted across multiple dimensions, encompassing the quality of individual frames (PSNR and SSIM), content consistency (ACD and ACD-I), and video quality (FVD). The comprehensive quantitative and qualitative outcomes of our evaluation are presented in Fig.\ref{example-gen}, Fig.\ref{example-enh}, and Table.\ref{tab:results}.

\vspace{0.3cm}

\textbf{Qualitative Results:} 

In Fig. \ref{example-gen}, we present illustrative examples that underscore the efficacy of our proposed model in enhancing facial expression videos. The figure showcases instances of various identities expressing different facial expressions. On the left side, we present input image identities, while in the middle, we display the same identities performing one of the six basic facial expressions. Additionally, on the right side, we depict a single individual executing the six basic facial expressions, with each expression represented in a separate line. This visual representation effectively demonstrates the successful generation of expressions in high-resolution videos, highlighting the model capability to enhance facial expressions across diverse expressions and identities. Additional examples of videos generated by our model are presented in Fig.\ref{fig:Apndx1} in the Appendix section.
\vspace{0.2cm}

Moreover, in Fig. \ref{example-enh}, we present videos generated by low-resolution models: VideoGAN, ImaGINator, MotionGAN, and FEV-GAN, alongside their corresponding enhanced videos produced by our model. The figure visually captures the noticeable improvements in video quality, background representation, and enhanced details such as hair and clothing, contributing to an overall enhancement in visual fidelity. Beyond these enhancements, our model exhibits additional valuable properties. For instance, when VideoGAN (a) encounters challenges in identity preservation, resulting in identity loss during the generation of expressions, our enhancement (b) successfully recovers facial identity details. Furthermore, our model ensures clarity in identity representation by incorporating details such as hair and clothing. When addressing ImaGINator (c), which generates expressions that are barely noticeable, blurred, and noised, our model (d) effectively synthesizes clear expressions while eliminating noise and blurriness. In the case of MotionGAN (e), which introduces distortions in the mouth and eyes areas, impacting the smoothness of the video, our enhancement (f) rectifies these distortions. As for FEV-GAN, which exhibits minimal distortions in low resolution, our model effortlessly enhances it to higher resolution.

\begin{table}[h]
\caption{Quantitative Results for Facial Expression Enhancement. The table presents results for videos generated by low-resolution models (VideoGAN, ImaGINator, MotionGAN, FEV-GAN) both before and after enhancement using our proposed model.}\label{tab:results}%
\begin{tabular}{@{}c|c|c|c|c|c|c @{}}
\toprule
\textbf{Model } & \textbf{Resolution}& \textbf{FVD $\downarrow$}& \textbf{PSNR $\uparrow$}&  \textbf{SSIM $\uparrow$}& \textbf{ACD $\downarrow$}& \textbf{ACD-I $\downarrow$}\\
\midrule
\textbf{VideoGAN} & (32x64x64) & 712.55 & 22.30 & 0.83 & 0.09 & 0.73\\
\textbf{FacEnhance$_{VideoGAN}$} & (32x192x192) & 91.31 & 30.88 & 0.94 & 0.01 & 0.09  \\
\hline
\hline
\textbf{ImaGINator} & (32x64x64) & 2205.26 & 20.29 & 0.85 & 0.08 & 0.29\\
\textbf{FacEnhance$_{ImaGINator}$} & (32x192x192) & 111.99 & 30.27 & 0.93 & 0.01 & 0.05 \\
\hline
\hline
\textbf{MotionGAN} & (32x64x64) & 654.98 & 23.96 & 0.89 & 0.011 & 0.29 \\
\textbf{FacEnhance$_{MotionGAN}$} & (32x192x192) & 167.01 & 31.08 & 0.95 &  0.010 & 0.11 \\
\hline
\hline
\textbf{FEV-GAN} & (32x64x64) & 452.46 & 24.02 & 0.88 & 0.09 & 0.23 \\
\textbf{FacEnhance$_{FEV-GAN}$} & (32x192x192) & 85.74 & 31.08 & 0.93 &  0.01 & 0.09 \\
\botrule
\end{tabular}
\end{table}

\textbf{Quantitative Results:}

Table \ref{tab:results} demonstrates the effectiveness of our model performance, showcasing its ability to effectively increase the resolution from (32x64x64) to (32x192x192), while improving the quality of the generated videos, as demonstrated by notable gaps in FVD, PSNR, and SSIM scores between the low-resolution generated videos and their enhanced counterparts. the model also preserves the smoothness of the expressions and the input identity as indicated in ACD and ACD-I values.
\vspace{0.2cm}

Both quantitative and qualitative analyses consistently affirm the effectiveness of our model in achieving substantial enhancements in facial expression videos. This is evident in the case of VideoGAN, which exhibits challenges in identity preservation, as reflected in ACD-I scores reaching as high as 0.73, compared to the lower scores of ImaGINator, MotionGAN, and FEV-GAN ranging from 0.2 to 0.3 for other models. However, our proposed model successfully addresses identity loss issue, achieving an ACD-I as low as 0.09. Comparable patterns are noted with ImaGINator, where the model produces high FVD 2205.26 and low PSNR 20.09, indicating temporal and spatial difficulties and very low video quality. With the application of our model, there is a significant improvement, reflected in the decreased FVD to 111.99 and increased PSNR to 30.27, indicating a significant enhancement in the quality of the generated videos. These results underscore the consistent and robust performance of our proposed model in overcoming identity preservation and quality challenges posed by existing models.

\subsubsection{Comparison To State-of-the-art Models}

\begin{figure}[t]
\begin{center}
\includegraphics[scale=0.45]{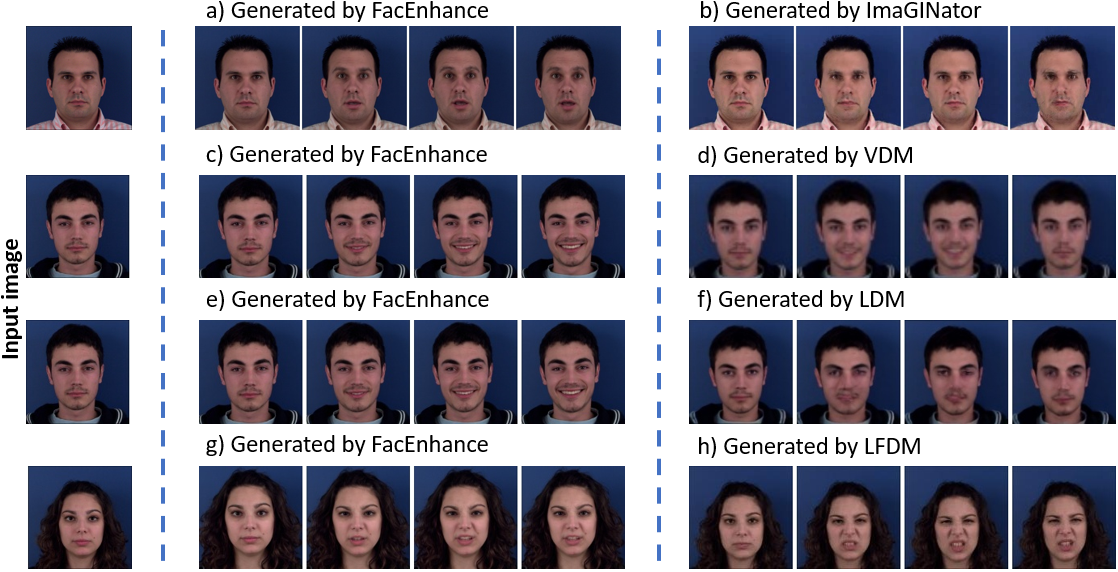}
\caption{Qualitative comparison of sequences generated by the FacEnhance model and state-of-the-art models on the MUG database. The sequences of our model (a, c, e, g), ImaGINator (b), and LFDM (f) are randomly selected from the test results, and the sequences of VDM (c) and FDM (e) are taken from \cite{ni2023conditional}.}
\label{sota-cmpr}
\end{center}
\end{figure}

In this section, we conduct a comprehensive comparison between our proposed model and state-of-the-art models for facial expression generation, including ImaGINator \cite{wang2020imaginator}, VDM \cite{ho2022video}, LDM \cite{rombach2022high}, and LFDM \cite{ni2023conditional}. 
\vspace{0.3cm}

\textbf{Qualitative Comparison}

Fig. \ref{sota-cmpr} visually presents facial expression videos generated by our model alongside those generated by the baselines. The left side of the figure displays the input identity, our generated videos (a, c, e, g) are showcased in the middle, and baseline videos (b, d, f, h) are presented on the right side. Each line in the comparison corresponds to videos generated by our model and a baseline, depicting the same identity and expression. It is important to note that the resolutions vary among the models (resolution details provided in Table \ref{tab:SOTA}). Despite these variations, both our model and the baselines exhibit robust identity preservation capabilities. Notably, ImaGINator (b), VDM (d), and LDM (f) show artifacts and blurriness in the mouth and nose area. In contrast, our proposed model consistently yields clearer expressions, improved facial details, and fewer distortions. When compared to LFDM (h), our proposed model demonstrates competitive results, highlighting its efficacy in enhancing facial expressions generated by low-resolution methods. 
\begin{table}[h]
\caption{ Quantitative comparison results of FacEnhance and baseline models.}\label{tab:SOTA}%
\begin{tabular}{@{}c|c|c|c|c|c|c @{}}
\toprule
\textbf{Model} &\textbf{Resolution} & \textbf{FVD $\downarrow$}& \textbf{PSNR $\uparrow$}&  \textbf{SSIM $\uparrow$}& \textbf{ACD $\downarrow$}& \textbf{ACD-I $\downarrow$}\\
\midrule
\textbf{ImaGINator (WACV 2020) \cite{wang2020imaginator}} & (32x192x192) & 415.53 & 22.32 & 0.88 & 0.08 & 0.17 \\
\textbf{VDM (ArXiv 2022)\cite{ho2022video}} & (16x64x64) & 108.02 & - & - & - & -  \\
\textbf{LDM (CVPR 2022) \cite{rombach2022high}} & (40x128x128) & 126.28 & - & - & - & -\\
\textbf{LFDM (CVPR 2023) \cite{ni2023conditional}} & (40x128x128) & 117.24 & 22.94 & 0.79 & 0.04 & 0.20 \\
\textbf{FacEnhance$_{MUG}$} & (32x192x192) & \textbf{95.38} & \textbf{30.88} &  \textbf{0.94} & \textbf{0.01} & \textbf{0.07}   \\
\botrule
\end{tabular}
\end{table}
\vspace{0.3cm}

\textbf{Quantitative Comparison}

Table \ref{tab:SOTA} provides a comprehensive quantitative comparison between our proposed model and baseline methods. FacEnhance demonstrates superior reconstruction capabilities, as evidenced by lower Frechet Video Distance (FVD) scores, higher Peak Signal-to-Noise Ratio (PSNR) and Structural Similarity Index (SSIM) scores. The observed superiority of FacEnhance suggests its efficacity in producing higher-quality videos with reduced noise and preserved overall structure throughout the entire video sequence, even when comparing videos at higher resolutions (32x192x192) generated by FacEnhance against baseline resolutions (16x64x64 or 40x128x128). The evaluation of content consistency, as assessed by ACD, highlights the superior performance of FacEnhance compared to the baselines. This improvement is attributed to the incorporation of the previous frame $\Bar{f}_{high}^{n-1}$ in the frame generation process, enhancing the temporal context of the video and improving content consistency. In terms of identity preservation, our proposed model demonstrates a significant performance advantage over ImaGinator and LFDM, indicating its efficacy in preserving facial identities throughout the enhancement process. 
\vspace{0.2cm}

The coherence observed in both qualitative and quantitative results emphasizes the ability of the proposed model in enhancing facial expression videos not only in terms of visual quality and content consistency but also in maintaining the integrity of facial identities.

\subsubsection{Ablation Study:}
\label{sec:ablation}
In this section, our focus is on illustrating the significance of the principal components employed in constructing our model. We specifically highlight the impact of the low-resolution conditioning expression embedding $E_{exp}$, the use of the previous frame $\Bar{f}_{high}^{n-1}$, and the addition of noise to the previous frame ($\Bar{f}_{high}^{n-1}+z$). This investigation is carried out through an ablation study, involving multiple versions of our model, in which we selectively eliminate certain components to observe their effects. Three new model versions are trained for this purpose. The first version involves discarding the expression encoder $E_{exp}$ and inputting the expression into the model concatenated with the remaining input images. In the second version, we exclude the use of the previous frame $\Bar{f}_{high}^{n-1}$ in the frame generation process. In the thirst version, we train the model using the previous frame $\Bar{f}_{high}^{n-1}$, but without adding noise to it ($\Bar{f}_{high}^{n-1}+z$). Then, we compare the performance of these ablated versions with the full proposed model. All four networks undergo training and evaluation using the same dataset, parameters, losses, and number of epochs.
\begin{table}[h]
\caption{Performance comparison of the full model against variants that exclude the expression encoder, previous frame (t-1), and the noisy previous frame in the generation process. }\label{tab:ablation}%
\begin{tabular}{@{}c|c|c|c|c|c @{}}
\toprule
\textbf{Model} & \textbf{FVD $\downarrow$}& \textbf{PSNR $\uparrow$}&  \textbf{SSIM $\uparrow$}& \textbf{ACD $\downarrow$}& \textbf{ACD-I $\downarrow$}\\
\midrule
\textbf{ W/o exp encoder} & 110.13 & 30.97 & 0.93 & 0.01 & 0.08 \\
\textbf{W/o previous frame (t-1)} & 126.60 & 31.55 & 0.94 & 0.01 &  0.06\\
\textbf{W/o noisy (t-1)} & 721.03 & 22.35 & 0.72 & 0.01 & 0.10\\
\textbf{Full FacEnhance} & 95.38 & 30.88 & 0.94 & 0.01 & 0.07  \\
\botrule
\end{tabular}
\end{table}

The results of our ablation study are presented in Table \ref{tab:ablation}. Comparing our full model to the model without the expression encoder $E_{exp}$ reveals closely matched PSNR, SSIM, ACD, and ACD-I values, suggesting comparable pixel-wise and structure-wise quality performance, as well as proficient handling of facial identity. However, the FVD metric indicates that the model without the expression encoder exhibits inferior video results, signifying a decrease in dynamic and temporal visual quality. This underscores the significant role played by the expression encoder, responsible for injecting the expression embedding into the model embedding space.
\vspace{0.2cm}

Regarding the incorporation of the previous frame in the generation process, training the model without $\Bar{f}_{high}^{n-1}$ yields frames with good quality and effective facial identity handling, as suggested by the PSNR and SSIM values, but diminished temporal dynamics and overall video distribution, as proved by the FVD. Adding $\Bar{f}_{high}^{n-1}$ without introducing noise results in even poorer visual and dynamic quality, as the model tends to overfit and overly depend on specific details of $\Bar{f}_{high}^{n-1}$, leading to frames with accumulated noise and distortions from the previous ones. However, with the addition of noise ($\Bar{f}_{high}^{n-1}+z$), the image loses detailed features but retains the silhouette and component positions. This introduces difficulties in overfitting on $\Bar{f}_{high}^{n-1}$, making detail generation more reliant on other images ($I_{Id}$ and $f^n_{low}$). The observed disparities between the full model and the ablated versions underscore the superior performance of the full model in capturing the overall video and motion structure, affirming the significance of the components employed in the proposed model.

\subsection{Discussion \& Limitations}

Our proposed model showcases robust capabilities in enhancing video content, effectively elevating resolution, improving overall quality, and incorporating complex details, including clothes, hair style, and background. A notable strength lies in its adept preservation of motion characteristics from low-resolution videos, successfully translating them into higher-quality outputs. In direct comparison with Video Diffusion Models (VDM), our model demonstrates heightened efficiency and outperforms alternative approaches, such as LFDM, particularly in terms of content consistency.

\begin{figure}[h]
\begin{center}
\includegraphics[width=240px]{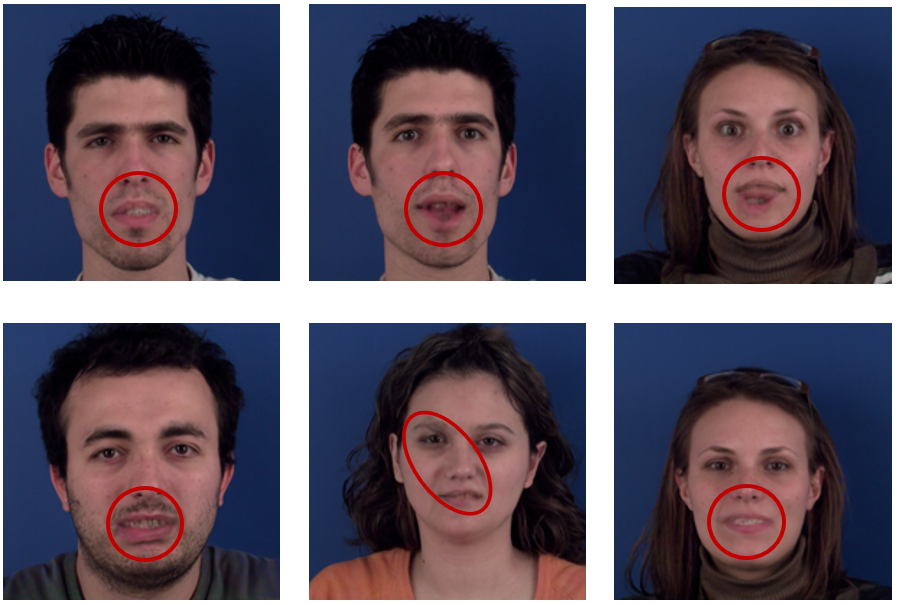}
\caption{Examples of flawed frames from videos enhanced by our model.}
\label{fig:limitations}
\end{center}
\end{figure}

\vspace{0.2cm}
Despite the notable successes of our model, it is essential to address inherent limitations and challenges. Occasional failures and distortions may manifest, particularly when the input low-resolution video comprises blurry or distorted images. The challenges become evident in scenarios where the visual information is inherently ambiguous or when the input quality deviates significantly from the training data. Figure \ref{fig:limitations} showcases instances of such occasional failures, highlighting the need for further refinement. Moreover, while our model demonstrates efficiency in comparison to methods such as VDM, it is crucial to recognize that the computational demands of our model remain substantial. The complexity of the underlying neural network architecture and of the high-dimensional facial data, may pose constraints on real-time applicability. 
\vspace{0.2cm}

\begin{figure}[h]
\begin{center}
\hspace{0.1cm} \includegraphics[width=240px]{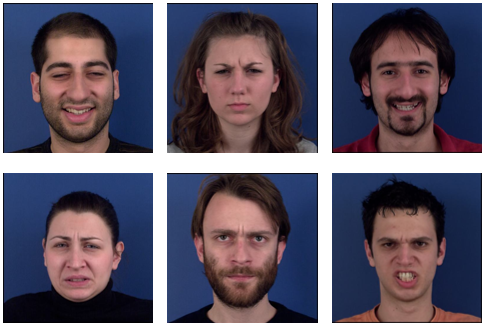}
\caption{Examples of our validation results at 512x512.}
\label{fig:512x512}
\end{center}
\end{figure}

As we envision future improvements, the proposed model holds the potential to achieve even higher resolutions, exemplified by our validation results at 512x512. Examples of validation results at 512x512 are illustrated in Fig.\ref{fig:512x512}. However, escalating resolution introduces exponential complexity, posing challenges in optimization, generalization, and mitigating overfitting. Additionally, the associated increase in training time presents a considerable obstacle.

\section{Conclusion \& Perspectives}\label{conclusion}

In conclusion, this paper introduced a novel framework for dynamic 2D facial expression enhancement, transforming low-resolution (64x64) facial videos without background to higher-resolution (192x192) with enhanced quality and added background details. Leveraging generated data from low-resolution expression generation models to guide a diffusion model, our approach successfully generates videos with improved quality while preserving the original motion characteristics.
\vspace{0.2cm}

Our experimental results showcased the effectiveness of the proposed enhancement method, demonstrating its competitiveness with the state-of-the-art. Despite its success, we acknowledge certain limitations, notably occasional failures and distortions that may arise during the enhancement process. Additionally, the computational demands of our approach currently inhibit real-time applicability.
\vspace{0.2cm}

To address these challenges, future work will focus on minimizing occasional failures and distortions by exposing the model to more diverse data through advanced data augmentation techniques. Furthermore, we aim to develop more efficient approaches to diffusion models, such as exploring the potential of LDMs, with the goal of real-time applicability and extending the scope to even higher resolutions.

\section*{Declarations}

\textbf{Conflict of interest/Competing interests:} The authors have no conflicts of interest to declare relevant to this article's content.

\vspace{0.5cm}
\noindent \textbf{Data availability:} The data used in this study was obtained from an external (public) source, and is not generated by the authors. Interested readers can the database through the following link: \hyperlink{https://mug.ee.auth.gr/fed/}{https://mug.ee.auth.gr/fed/}.

\begin{appendices}

\section{Extra results of FacEnhance}\label{secA1}
\begin{figure}
    \includegraphics[width=.48\textwidth]{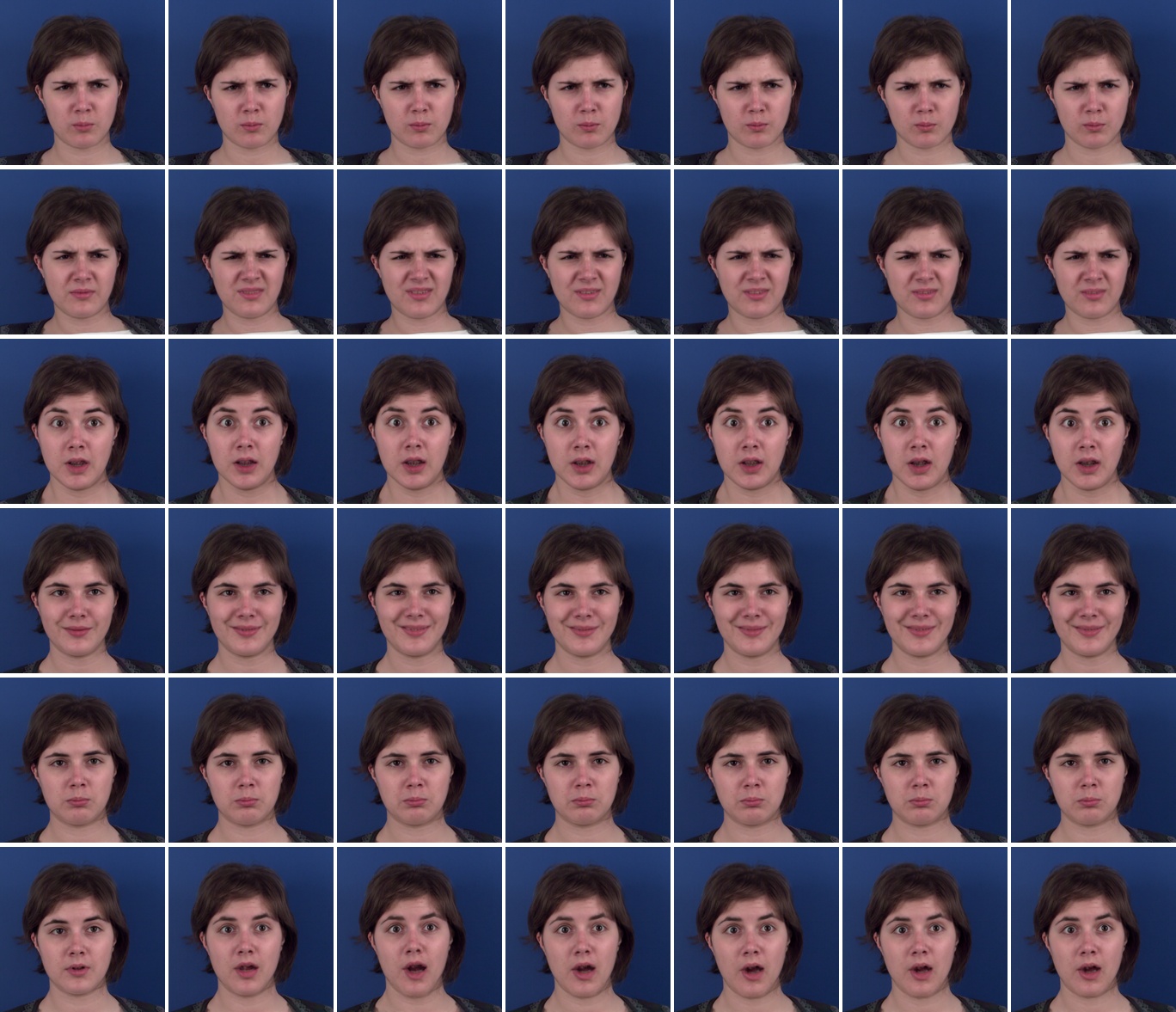}
    \includegraphics[width=.48\textwidth]{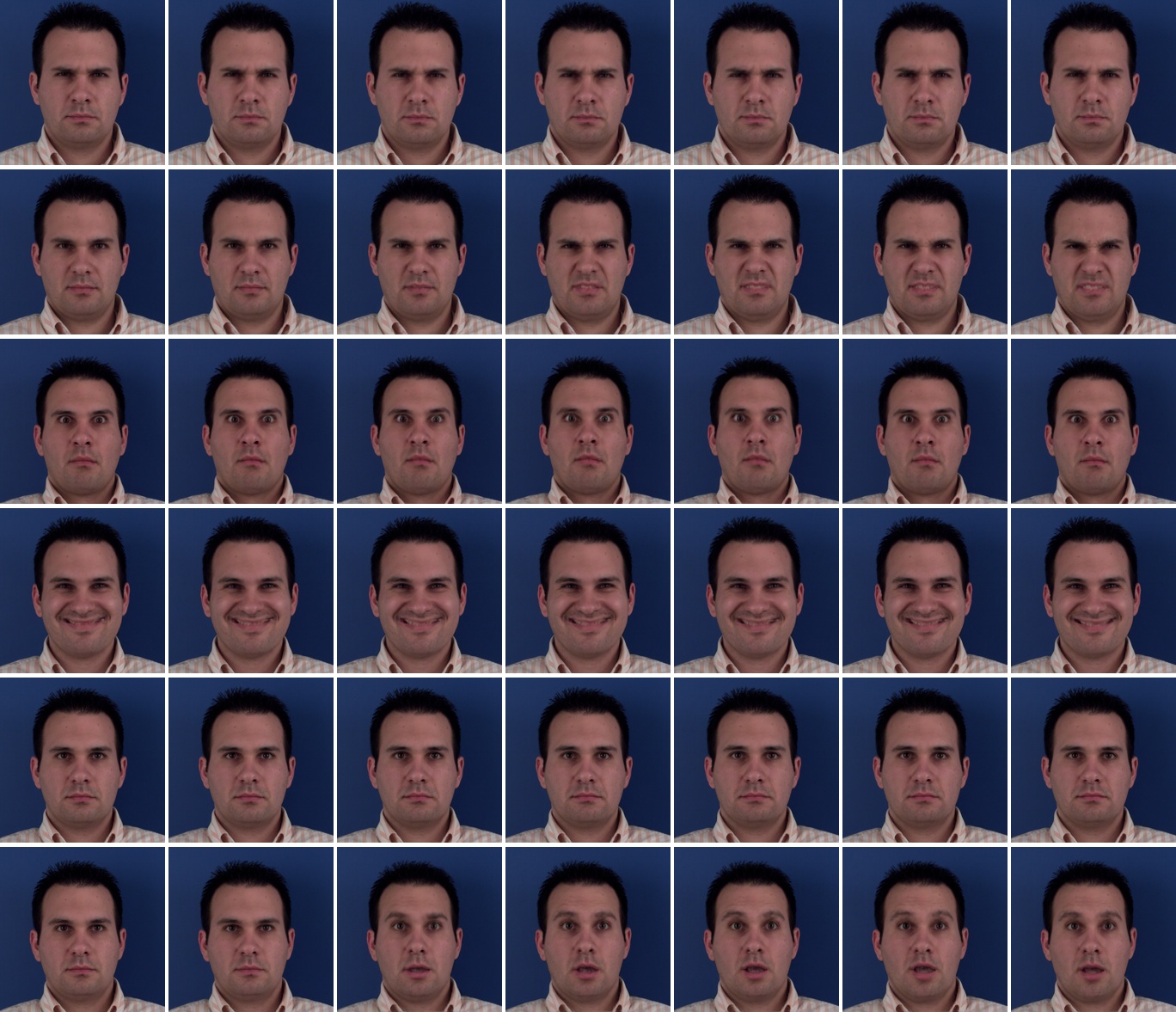}\\
    \includegraphics[width=.48\textwidth]{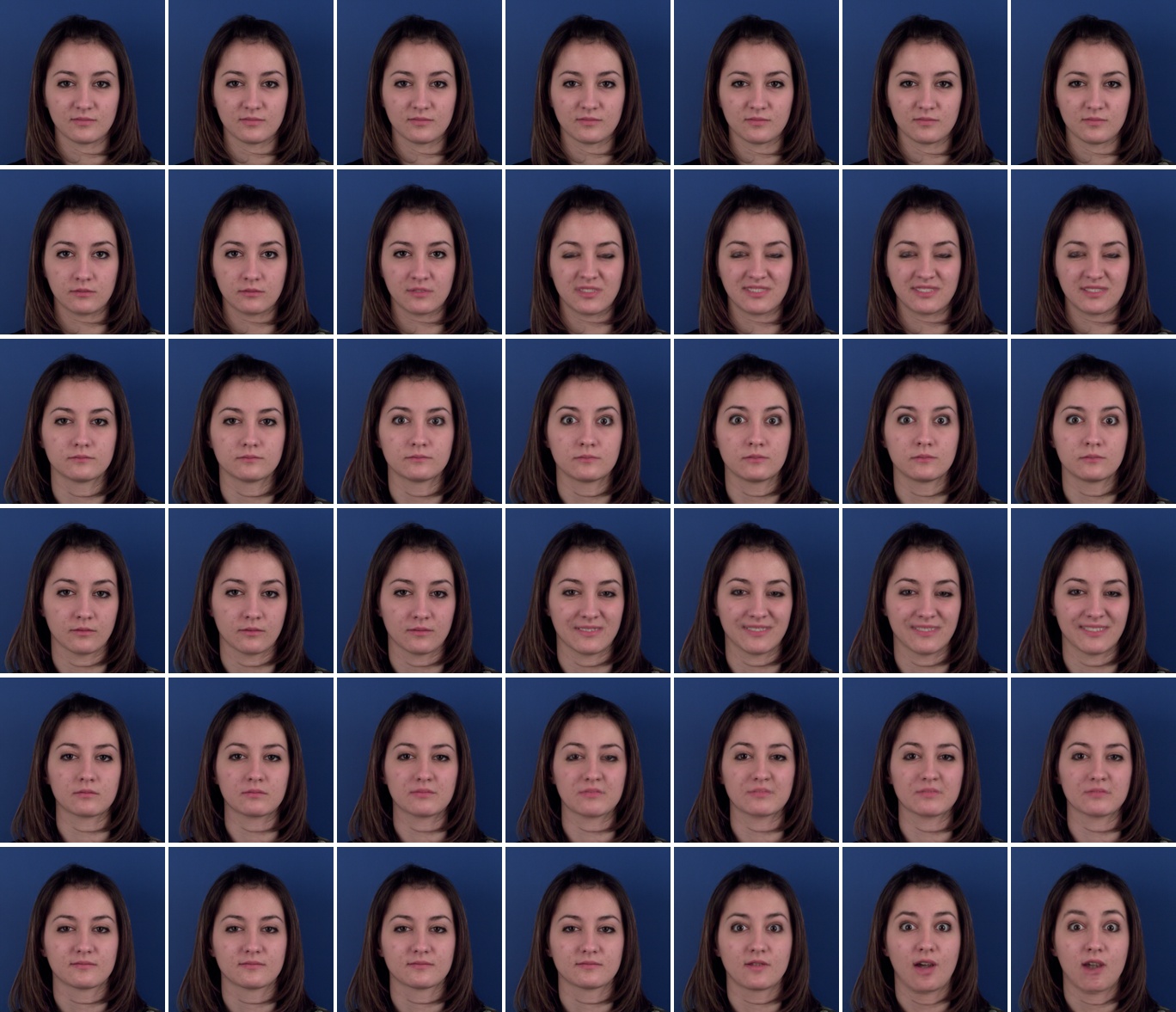}
    \includegraphics[width=.48\textwidth]{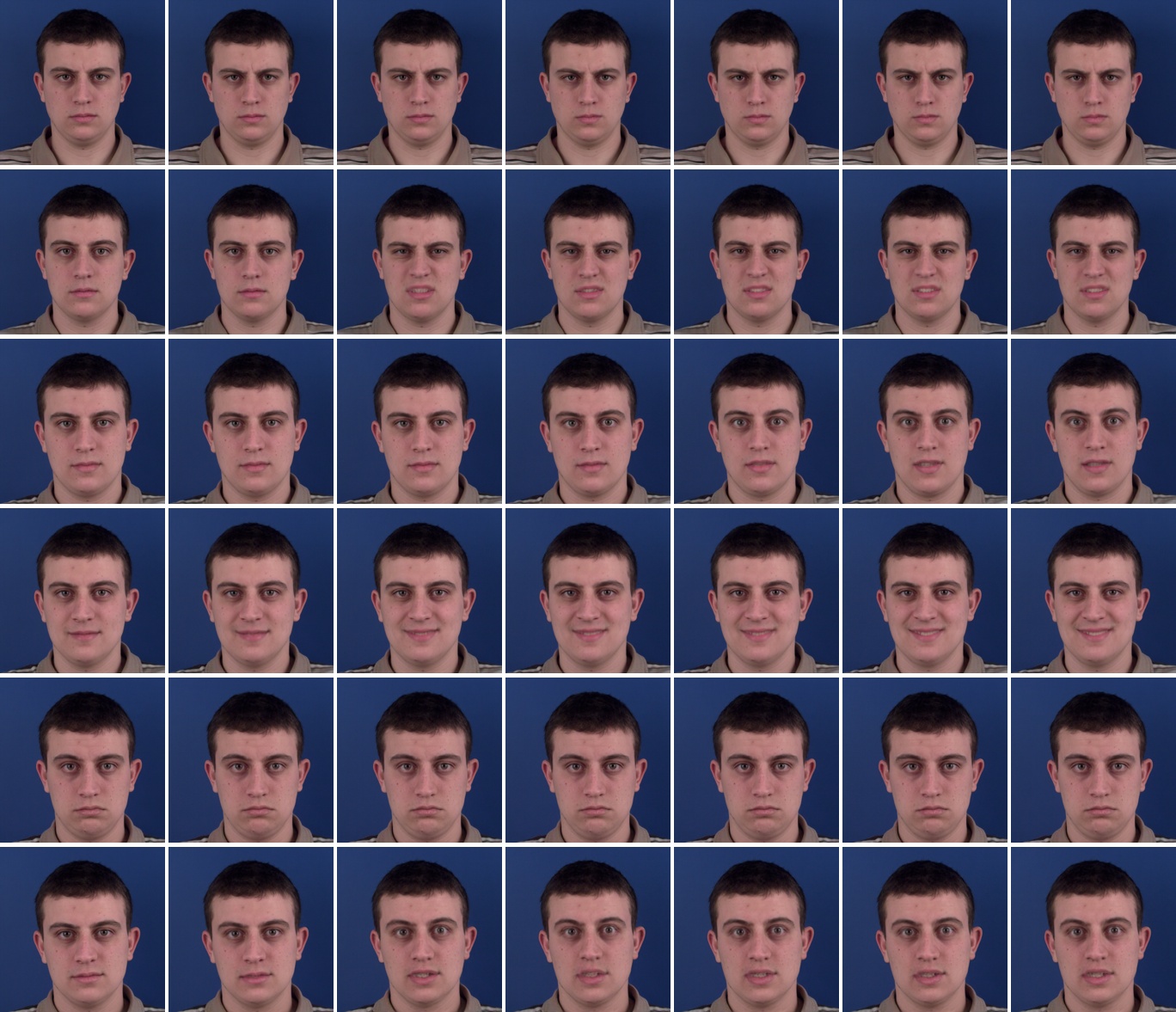}\\
    \includegraphics[width=.48\textwidth]{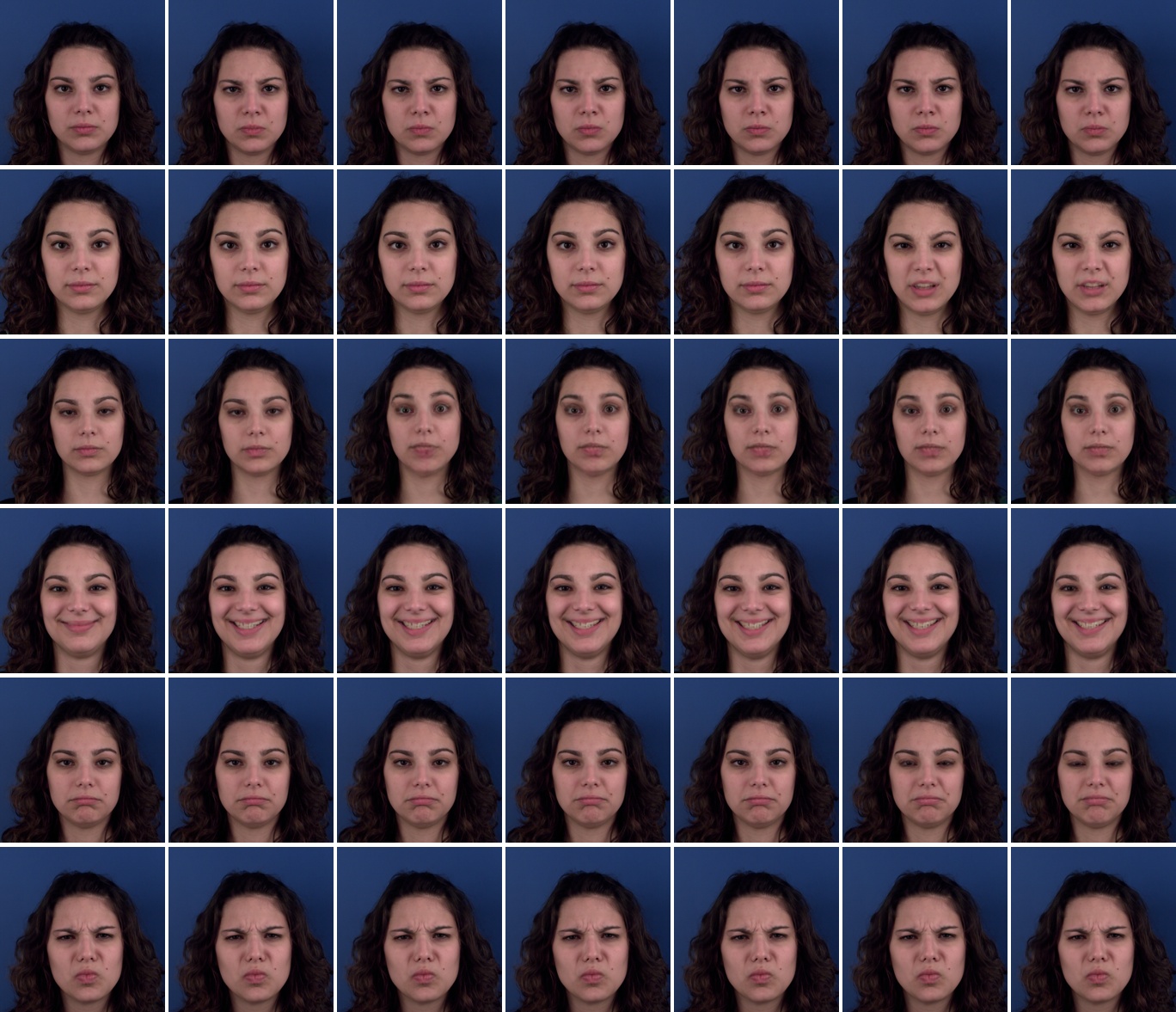}
    \includegraphics[width=.48\textwidth]{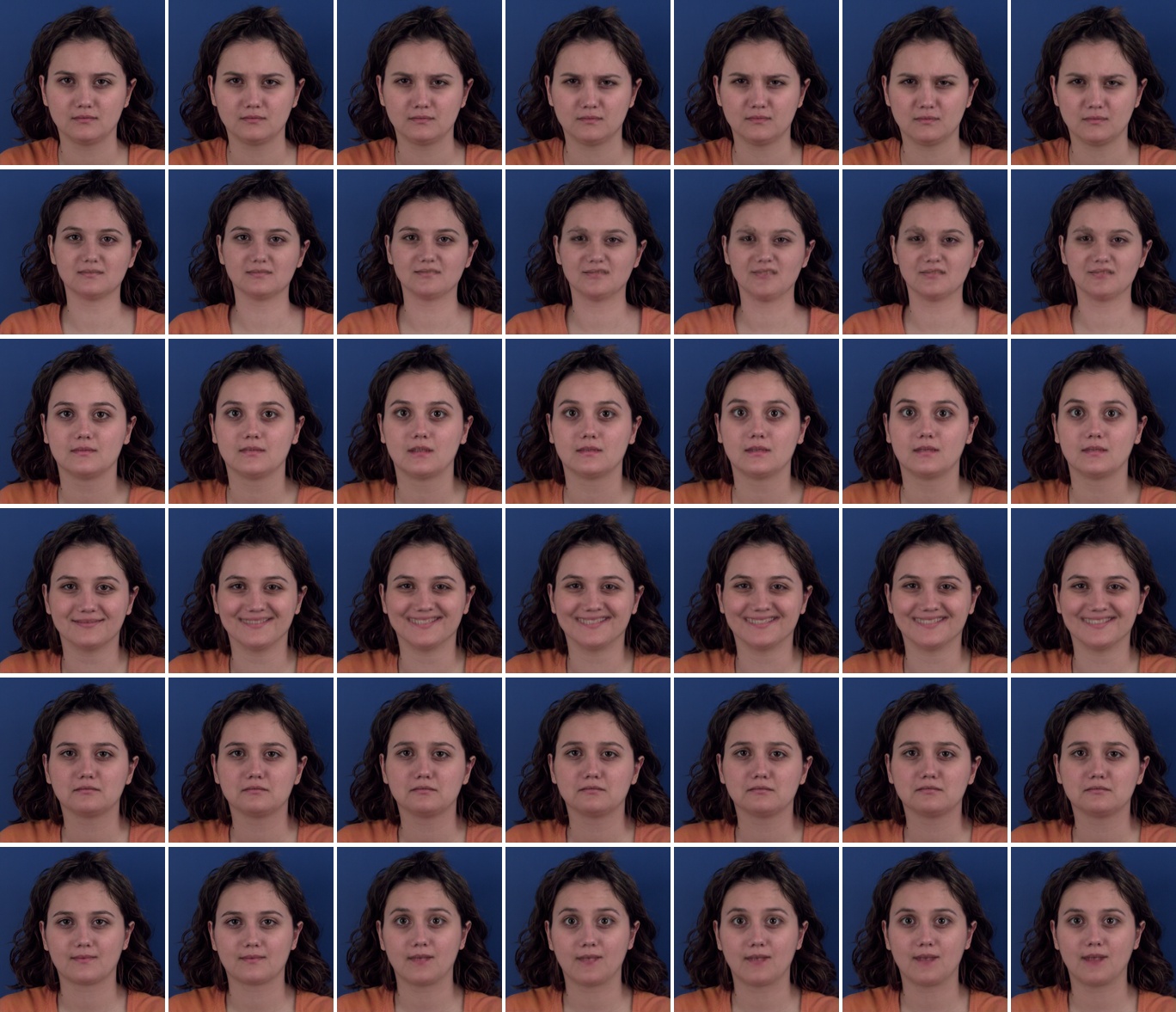}
    \caption{Examples of videos enhanced by our proposed model showcasing diverse identities performing the six basic facial expressions. These visual examples offer insights into the model performance and highlight its efficacy in enhancing realistic and diverse facial animations.}\label{fig:Apndx1}
\end{figure}



\end{appendices}

\vspace{-0.2cm}
\bibliography{sn-bibliography}

\end{document}